\documentclass[10pt,letterpaper]{article}
\usepackage[preprint]{tmlr}

\usepackage[utf8]{inputenc}
\usepackage[T1]{fontenc}
\usepackage{lmodern}
\usepackage{hyperref}
\usepackage{url}
\usepackage{booktabs}
\usepackage{amsfonts}
\usepackage{nicefrac}
\usepackage{microtype}
\usepackage{xcolor}
\usepackage{graphicx}
\usepackage{multirow}
\usepackage{amsmath}

\usepackage{natbib}

\definecolor{C0}{HTML}{3182ce}
\hypersetup{
  breaklinks=true,
  colorlinks=true,
  bookmarks=false,
  citecolor=C0,
  linkcolor=C0,
  urlcolor=C0
}

\title{SoU-RHS: Bayesian Structured Sparsity \\
for Compact Architecture Recovery in CNNs
}

\author{%
\name Rafael Mouallem Rosa \email rafael.mouallem-rosa@inria.fr \\
\addr Univ. Grenoble Alpes, Inria, CNRS, Grenoble INP, LJK,\\
38000 Grenoble, France
\AND
\name Julyan Arbel \email julyan.arbel@inria.fr \\
\addr Univ. Grenoble Alpes, Inria, CNRS, Grenoble INP, LJK,\\
38000 Grenoble, France
}

\begin{document}

\maketitle

\begin{abstract}
Highly parameterized convolutional neural networks often achieve strong predictive performance, but their memory, computation, and deployment costs motivate the search for methods that recover compact architectures from large reference models. We propose SoU-RHS, a Bayesian structured-sparsity procedure that augments Regularized Horseshoe Shrinkage (RHS) with the so-called Sources of Uncertainty Sparse Global-Local mechanism (SoU-SGL). The resulting model sharpens the separation between active and inactive filters, providing an effective mechanism for architecture recovery. 
A key property of SoU-RHS is stability: across different overparameterized starting widths and under a reasonable range of pruning thresholds, it recovers compact architectures with similar total size and layer-wise allocation.
This supports the interpretation that the procedure is identifying a compact structure adapted to the data, rather than merely compressing each initial model separately. 
Across standard image-classification benchmarks,
SoU-RHS recovers substantially smaller architectures than deterministic structured-sparsity baselines, 
while maintaining competitive accuracy, achieving lower negative log-likelihood in every evaluated configuration and generally lower expected calibration error.
\end{abstract}

\section{Introduction}

\paragraph{Recovering compact architectures.} Deep neural networks are often trained in highly parameterized regimes. Although such models can achieve strong predictive performance, they also incur substantial costs in memory, computation, and deployment. As machine learning systems are used at increasing scale, these costs are no longer secondary considerations, which has intensified the search for models that are not only accurate but also compact and efficient.

In many cases, such highly parameterized models are effectively overparameterized for the task at hand. An appealing alternative to manual architecture design is therefore to recover a compact architecture from a large reference model. Rather than postulating a small network a priori, one may start from a large model and let the data determine which structural components are actually needed.

\paragraph{Structured sparsity for architectural simplification.} This naturally brings structured sparsity into focus. Not all forms of sparsity lead to meaningful reductions in computational cost: unstructured sparsity removes individual weights, but often leaves the overall architecture unchanged. By contrast, structured sparsity removes entire architectural components, thereby simplifying the network itself. It is this architectural simplification, rather than sparsity per se, that most directly translates into computational savings.

\paragraph{From Bayesian shrinkage to robust filter selection.} In the convolutional setting considered here, compact architecture recovery is most directly posed as deciding which filters to retain. A natural Bayesian approach is then to place a horseshoe-type shrinkage prior on filter-wise components. The horseshoe prior was originally introduced for sparse-signal settings \citep{carvalho2010horseshoe}, encoding the prior belief that many coefficients are negligible while a few strong signals should remain active. In our setting, the same logic is transferred to filters: some filters are expected to be essential, while others should be shrunk toward inactivity. 

However, shrinkage alone does not necessarily produce sparsity that is sufficiently pronounced for robust pruning. Horseshoe shrinkage can strongly reduce weakly supported filters while still leaving them at small nonzero magnitudes.
The Sources of Uncertainty Sparse Global-Local mechanism (SoU-SGL) \citep{rosa2026bayesian} addresses this by selectively driving the local shrinkage scales toward an optimization-like treatment, producing a clearer separation between active and inactive filters. 
This is essential for compact architecture recovery, since the goal is not only to reduce filter magnitudes but also to robustly identify which filters should remain in the final architecture.

\paragraph{Contributions.}
We propose SoU-RHS, a Bayesian structured-sparsity approach for compact architecture recovery in convolutional neural networks (CNNs). SoU-RHS combines filter-wise regularized horseshoe shrinkage with SoU-SGL:
\begin{itemize}
    \item It produces pruning scores that more clearly separate inactive and active filters.
    \item Across standard image-classification benchmarks, it recovers substantially smaller architectures than deterministic structured-sparsity baselines, and its recovered widths are more stable across pruning thresholds and overparameterized starting widths.
    \item These compact architectures maintain competitive accuracy relative to the baselines while 
    achieving consistently lower negative log-likelihood and generally lower
expected calibration error than the deterministic baselines.
    \item Retraining on the recovered architectures further reveals an architecture--inference interaction, with deterministic methods deteriorating while Bayesian models remain strong or improve.
\end{itemize}

\paragraph{Outline.}
The remainder of the paper is organised as follows. Section~\ref{sec:rw} reviews related work. 
Section~\ref{sec:Methods} presents the baseline models and introduces SoU-RHS and its filter-level pruning mechanism. 
Section~\ref{sec:expe} describes the experimental setup, and Section~\ref{sec:res} presents results on architecture recovery, predictive performance, and retraining. Section~\ref{sec:lim} discusses limitations, and Section~\ref{sec:conc} concludes. Appendix~\ref{sec:post_sampling} examines the trade-off between posterior sampling and predictive cost.

\section{Related work\label{sec:rw}}

\paragraph{Deterministic structured sparsity in CNNs.}
Within CNNs, structured sparsity is often operationalized by pruning filters or channels, thereby turning sparsity into an explicit change in architecture.
One line of work follows a train--prune--fine-tune pipeline, removing filters or channels according to magnitude, reconstruction, or saliency criteria \citep{li2017pruning,he2017channel,luo2017thinet,molchanov2017pruning}. Another approach induces structured sparsity during training through explicit regularization. Structured Sparsity Learning (SSL) uses group-lasso penalties over predefined structures such as filters and channels \citep{wen2016learning}, while Network Slimming promotes channel sparsity through an $\ell_1$ penalty on batch-normalization scaling parameters \citep{liu2017learning}. Related approaches such as MorphNet and AutoSlim formulate channel-width selection as resource-constrained structure learning or architecture search \citep{gordon2018morphnet,yu2019autoslim}.
More recent work addresses dependencies between structures across layers. DepGraph \citep{fang2023depgraph} groups coupled parameters for consistent structural pruning across diverse architectures. 
Such dependencies are also relevant when extending filter-wise shrinkage to networks with residual connections.

\paragraph{Bayesian structured sparsity.}
Bayesian approaches to structured sparsity differ mainly in where the sparsity-inducing random variables enter the network. One line places dropout-like multiplicative noise directly on structural units, such as neurons or channels, and prunes units whose scales collapse toward zero \citep{neklyudov2017structured,wright2024bmrs}. Another line places global-local shrinkage priors on groups of weights associated with structural units, allowing entire weight groups to be shrunk toward inactivity \citep{louizos2017bayesian,ghosh2018structured,ghosh2019model}. 
Our work follows this second line: we apply regularized horseshoe shrinkage to filter-wise weight groups in CNNs and use SoU-SGL to sharpen the separation between active and inactive filters.

Spike-and-slab approaches make structural selection explicit. \citet{jantre2024spike} combine group Lasso and group horseshoe shrinkage with spike-and-slab priors to induce node sparsity in Bayesian neural networks. Our approach instead retains continuous filter-wise shrinkage and sharpens structural decisions through the inferential treatment of local scales.

Recent Bayesian methods also connect pruning to model evidence. BMRS \citep{wright2024bmrs} combines multiplicative-noise models with Bayesian model reduction, including a variant that avoids pruning-threshold tuning. SpaM \citep{Dhahri2024} uses marginal-likelihood optimization to encourage sparsifiability and derives a pruning criterion from Laplace posterior precision. 
\citet{Tazawa2026} connect synaptic pruning with online Bayesian model selection, using connectivity parameters to remove uninformative connections in generative models. Their work shares our view of pruning as statistical structure learning, while addressing individual connections rather than filter-level architecture recovery in CNNs. 
These approaches provide relevant alternatives to our use of local-scale uncertainty to stabilize filter selection.

\paragraph{Pruning as architecture selection.}
\citet{liu2019rethinking} show that, in their structured-pruning experiments, training recovered architectures from scratch can match or outperform fine-tuning inherited weights. This motivates treating pruning as architecture selection and evaluating the recovered structure independently of its learned weights. Our retraining study further examines how the value of a recovered architecture depends on the inference procedure.

\section{Structured Sparsity Models and SoU-RHS}
\label{sec:Methods}

Consider a convolutional neural network with $L$ layers and parameters
$\mathcal{W} = \{\mathcal{W}_\ell\}_{\ell=1}^L$.
For each convolutional layer $\ell$, write
$\mathcal{W}_\ell = \{W_{\ell c}\}_{c=1}^{C_\ell}$,
where $W_{\ell c}$ is the full weight tensor of filter $c$.
Given an input $x$, the network defines a predictor $f(x;\mathcal{W})$.
For a training dataset $\mathcal{D}=\{(x_n,y_n)\}_{n=1}^N$, we use the empirical classification loss

\[
\mathcal{L}_{\mathrm{data}}(\mathcal{W})
=
\frac{1}{N}\sum_{n=1}^N
\mathrm{CE}\!\left(y_n, f(x_n;\mathcal{W})\right),
\]
where \(\mathrm{CE}(\cdot,\cdot)\) denotes the cross-entropy loss.

We compare deterministic and Bayesian models, including dense baselines and methods that induce filter-level sparsity. 
For the sparsity-inducing methods, pruning is driven by a filter-specific quantity, which we refer to as the pruning score.
When this quantity is close to zero, the corresponding filter can be removed.
This provides a common language for both deterministic and Bayesian models.

\subsection{Deterministic neural networks}

A standard deterministic neural network (NN) is obtained by minimizing only the data-fitting term,
\[
\min_{\mathcal{W}} \mathcal{L}_{\mathrm{data}}(\mathcal{W}).
\]
To induce filter-level sparsity, this objective can be augmented with an additional regularization term,
\[
\min_{\mathcal{W},\,\gamma}
\mathcal{L}_{\mathrm{data}}(\mathcal{W},\gamma)
+
\mathcal{R}(\mathcal{W},\gamma),
\]
where $\gamma$ denotes any auxiliary filter-specific parameters introduced by the method.
Different deterministic pruning methods correspond to different choices of $\mathcal{R}$ and to different definitions of the pruning score driven toward zero.

\paragraph{Structured Sparsity Learning (SSL).}
In SSL \citep{wen2016learning}, filter-level sparsity is induced through a group-lasso penalty,
\[
\min_{\mathcal{W}}
\mathcal{L}_{\mathrm{data}}(\mathcal{W})
+
\lambda
\sum_{\ell=1}^L \sum_{c=1}^{C_\ell}
\|W_{\ell c}\|_2.
\]
Here the full weight tensor of each filter is treated as a group.
When $\|W_{\ell c}\|_2$ approaches zero, the corresponding filter is effectively removed.

\paragraph{Network Slimming (SLIM).}
Network Slimming \citep{liu2017learning} induces filter-level sparsity through the scaling parameters of batch normalization.
Let $\gamma_{\ell c}$ denote the scaling factor associated with filter $c$ in layer $\ell$.
The objective is
\[
\min_{\mathcal{W},\gamma}
\mathcal{L}_{\mathrm{data}}(\mathcal{W},\gamma)
+
\lambda \sum_{\ell=1}^L \sum_{c=1}^{C_\ell}
|\gamma_{\ell c}|.
\]
When $|\gamma_{\ell c}|$ approaches zero, the corresponding filter becomes inactive and can be pruned.

\subsection{Bayesian neural networks}

In a Bayesian neural network (BNN) \citep{neal1996bayesian,arbel2024primer,agostinelli2026BDL}, the weights are treated as random variables.
Given a prior \(\pi(\mathcal{W})\) and likelihood \(p(y_n \mid f(x_n;\mathcal{W}))\), inference is based on the posterior
\[
\pi(\mathcal{W}\mid \mathcal{D})
\propto
\pi(\mathcal{W})
\prod_{n=1}^N
p(y_n \mid f(x_n;\mathcal{W})).
\]
We approximate this posterior with a variational distribution \(q(\mathcal{W})\) that minimizes the variational objective \citep{blundell2015weight}
\begin{equation}\label{eq:ELBO}
\mathcal{J}(q)
=
\mathbb{E}_{q(\mathcal{W})}
\!\left[\mathcal{L}_{\mathrm{data}}(\mathcal{W})\right]
+
\frac{1}{N}\mathrm{KL}\!\left(q(\mathcal{W})\,\|\,\pi(\mathcal{W})\right).    
\end{equation}
This yields the usual BNN objective: an expected data-fitting term plus a KL penalty. Under an isotropic Gaussian prior on each filter tensor,
\[
W_{\ell c} \sim \mathcal{N}(0,\sigma_\ell^2 I),
\]
we obtain a baseline BNN.
We use a mean-field Gaussian variational family for the weights in this baseline BNN.
As with the deterministic NN baseline, this model does not explicitly induce filter-level sparsity.

\paragraph{Horseshoe prior for filter pruning.}
To induce filter-level sparsity in the Bayesian setting, we replace the Gaussian prior of the baseline BNN with a horseshoe prior on each filter tensor \citep{ghosh2018structured,ghosh2019model}. For filter $c$ in layer $\ell$, we write
\[
W_{\ell c} \mid \tau_{\ell c}, \nu_\ell
\sim
\mathcal{N}\!\left(0,\, \tau_{\ell c}^2 \nu_\ell^2 I\right),
\qquad
\tau_{\ell c} \sim \mathcal{C}^+(0,b_0),
\qquad
\nu_\ell \sim \mathcal{C}^+(0,b_g),
\]
where $\tau_{\ell c}$ is a filter-specific local scale and $\nu_\ell$ is a layer-wise global scale. The global scale $\nu_\ell$ controls the overall amount of shrinkage in layer $\ell$, while the local scale $\tau_{\ell c}$ allows individual filters to escape this shrinkage when supported by the data.

An equivalent non-centred parameterization is
\[
W_{\ell c} = s_{\ell c}\,\beta_{\ell c},
\qquad
s_{\ell c} = \tau_{\ell c}\nu_\ell,
\qquad
\beta_{\ell c} \sim \mathcal{N}(0,I).
\]
This representation makes the shrinkage mechanism explicit: the same scalar $s_{\ell c}$ multiplies all raw weights $\beta_{\ell c}$ in filter $c$.
Thus, when $s_{\ell c}$ approaches zero, the full filter is jointly shrunk toward zero and can be pruned.

In our implementation, we follow \citet{ghosh2018structured,ghosh2019model} and use the regularized horseshoe formulation \citep{Piironen2017}. The regularized horseshoe introduces an additional layer-wise parameter \(c_\ell^2\), which bounds the effective variance of filters that escape shrinkage. The raw global-local scale is replaced by the effective scale
\[
\tilde{s}_{\ell c}^2
=
\frac{c_\ell^2 s_{\ell c}^2}{c_\ell^2+s_{\ell c}^2}.
\]
The filter weights are then given by 
\(W_{\ell c}=\tilde{s}_{\ell c}\beta_{\ell c}\).
This preserves the strong shrinkage behavior of the horseshoe for weakly supported filters, while regularizing filters that escape the global-local shrinkage. We place an inverse-gamma prior on \(c_\ell^2\). 
This defines the regularized horseshoe model (RHS).
For architecture recovery, pruning remains based on the raw global-local scale \(s_{\ell c}\), which directly encodes the filter-wise shrinkage induced by the local and global scales.

\paragraph{Sources of Uncertainty Sparse Global-Local mechanism (SoU-SGL).}
We strengthen the regularized horseshoe formulation with SoU-SGL \citep{rosa2026bayesian}. The SoU framework extends Bayesian inference by partitioning the model’s unknown quantities into components, each representing a distinct source of uncertainty, and assigning each source \(k\) its own confidence parameter \(\gamma_k\). These parameters control the degree of probabilistic treatment assigned to each component. As \(\gamma_k\) decreases, the corresponding posterior factor becomes increasingly concentrated and the treatment of that component approaches optimization, while the remaining model components may retain a probabilistic treatment.

In the variational formulation, the confidence parameters enter as weights on the KL contributions associated with the different model components. For the global-local model considered here, using the non-centred parameterization above, the SoU KL term is
\[
\begin{aligned}
\mathcal{K}_{\mathrm{SoU}}(q)
={}&
\gamma_\beta
\sum_{\ell,c}
\mathrm{KL}\!\left(
q(\beta_{\ell c})
\,\|\,
\pi(\beta_{\ell c})
\right)
+
\gamma_\tau
\sum_{\ell,c}
\mathrm{KL}\!\left(
q(\tau_{\ell c})
\,\|\,
\pi(\tau_{\ell c})
\right)
\\
&+
\gamma_\nu
\sum_{\ell}
\mathrm{KL}\!\left(
q(\nu_\ell)
\,\|\,
\pi(\nu_\ell)
\right)
+
\gamma_c
\sum_{\ell}
\mathrm{KL}\!\left(
q(c_\ell^2)
\,\|\,
\pi(c_\ell^2)
\right).
\end{aligned}
\]
Setting all confidence parameters to one recovers the standard Bayesian variational objective, corresponding here to the regularized horseshoe model (RHS).

This decomposition is particularly useful here because the different model components play different roles: the local scales \(\tau_{\ell c}\) govern the filter-specific structural decision, whereas the remaining components govern the predictive model and its shrinkage hierarchy. SoU-SGL exploits this distinction by setting the local-scale confidence near zero, \(\gamma_\tau \approx 0\), while keeping \(\gamma_\beta=\gamma_\nu=\gamma_c=1\), corresponding to the standard Bayesian treatment for the remaining components. 
The local scales are therefore driven toward an optimization-like regime while the rest of the model remains probabilistic, yielding an “optimize structure, infer parameters” configuration. Applying this SoU-SGL specification to the regularized horseshoe defines our proposed model, SoU-RHS.

In SoU-RHS, the regularized horseshoe prior and the pruning score \(s_{\ell c}=\tau_{\ell c}\nu_\ell\) remain unchanged; what changes is the inferential treatment of the local scales. 
The near-zero confidence drives \(q(\tau_{\ell c})\) to concentrate more sharply around its data-driven solution.
For weakly supported filters, this limits posterior mass at larger local-scale values that would otherwise allow them to escape the global shrinkage, driving \(s_{\ell c}\) close to zero. Supported filters, in contrast, retain local scales large enough to escape the global shrinkage. 
This selective collapse of the local-scale posterior therefore yields a clearer separation between active and inactive filters, turning horseshoe shrinkage into a more decisive architecture-selection mechanism.

Unlike \(\ell_1\)-based methods such as SLIM and SSL, where \(\lambda\) directly tunes the strength of the sparsity penalty and therefore the compression--accuracy trade-off, \(\gamma_\tau\) is not used as a sparsity-strength parameter. 
We fix it near zero to sharpen the local-scale posteriors and the resulting filter-level structural decisions; the amount of pruning then emerges from the fitted global-local hierarchy.

For both RHS and SoU-RHS, we follow the fully factorized variational parameterization of \citet{ghosh2018structured,ghosh2019model}, using Gaussian factors for the non-centred weights \(\beta_{\ell c}\) and for the log-squared scale variables \(\log \tau_{\ell c}^2\), \(\log \nu_\ell^2\), and \(\log c_\ell^2\).

\section{Experiments\label{sec:expe}}

\paragraph{Datasets.}
We evaluate on four standard image-classification datasets: CIFAR-10, CIFAR-100, SVHN, and Tiny-ImageNet. CIFAR-10 and CIFAR-100 are 32$\times$32 natural-image datasets with 10 and 100 classes, respectively. SVHN is a 32$\times$32 real-world digit-image dataset with 10 classes. Tiny-ImageNet is a more challenging 64$\times$64 natural-image dataset with 200 classes. We use the standard training and test splits and do not use a separate validation split for checkpoint selection.

\paragraph{Architecture and overparameterized widths.}
All methods are implemented on the same VGG-style backbone with three convolutional blocks, each of the form
\[
\text{Conv} \rightarrow \text{ReLU} \rightarrow \text{Conv} \rightarrow \text{ReLU} \rightarrow \text{MaxPool}.
\]
All convolutions use \(3\times 3\) kernels with stride 1 and same padding. The final convolutional representation is followed by global average pooling and a linear classifier. We consider three overparameterized starting architectures, where all convolutional layers are assigned the same nominal width, equal to 128, 256, or 512 filters.

\paragraph{Baselines.} We compare SoU-RHS against four baselines in the main comparison: a standard deterministic neural network (NN), a Bayesian neural network (BNN) with a Gaussian prior, and the structured-sparsity methods Network Slimming (SLIM) and Structured Sparsity Learning (SSL), each evaluated with regularization strengths \(10^{-3}\) and \(10^{-4}\). 
We additionally include RHS in the stability analysis to isolate the effect of the SoU-SGL mechanism.
All methods use the same backbone.

\paragraph{Training and checkpointing.}
All models are optimized with Adam using batch size 128 and learning rate \(10^{-3}\). We do not use data augmentation or dropout. Deterministic models are trained for 500 epochs, whereas Bayesian models are trained for 4000 epochs. For Bayesian models, we use KL annealing: the KL coefficient starts at zero, increases linearly by \(5\times 10^{-4}\) per epoch, and is capped at \(0.5\). 
Normalization differs across model classes. Deterministic models use batch normalization, following standard CNN practice, whereas Bayesian models rely on global RMS normalization since channel-wise batch normalization would cancel the filter scale $s_{\ell c}$, which serves as the pruning score in RHS and SoU-RHS.

We set the regularized horseshoe hyperparameters to \(b_0=1\) for the local
half-Cauchy scale and \(b_g=10^{-1}\) for the global layer scale. For SoU-RHS,
following the SoU-SGL construction, we set \(\gamma_\tau=10^{-9}\) and keep
\(\gamma_\beta=\gamma_\nu=\gamma_c=1\).

For evaluation, we use the final checkpoint for Bayesian models. Unless otherwise stated, predictive statistics for Bayesian models are estimated by Monte Carlo averaging over 100 posterior samples. For deterministic models, we report the checkpoint with lowest training objective over the last 5\% of training epochs. This reduces sensitivity to the final iterate while avoiding a separate validation split.

\paragraph{Pruning rule and metrics.} 
For each sparsity-inducing method, pruning is based on its filter-level pruning score. 
For the deterministic structured-sparsity baselines, the pruning score is \(|\gamma_{\ell c}|\) for SLIM and \(\|W_{\ell c}\|_2\) for SSL. A filter is retained if its score exceeds a fixed threshold. In the main table, we use threshold \(10^{-2}\) for both deterministic methods. 
For SoU-RHS, the pruning score is the raw filter-wise global-local scale \(s_{\ell c}=\tau_{\ell c}\nu_\ell\). Since \(s_{\ell c}\) is random, a filter is pruned when
\[
\Pr(s_{\ell c} < \delta) > 0.9,
\]
where the probability is computed analytically under the variational posterior
of \(s_{\ell c}\). In the main table, we use \(\delta=10^{-3}\).

We report classification accuracy (Acc), negative log-likelihood (NLL), expected calibration error (ECE), and the number of active filters (\#AF). \#AF is counted after pruning. For sparsity-inducing methods, Acc, NLL, and ECE are evaluated after pruning, without additional fine-tuning. The NN and BNN baselines are evaluated without pruning.

\section{Results\label{sec:res}}

\begin{figure}[h]
    \centering
    \includegraphics[width=\columnwidth]{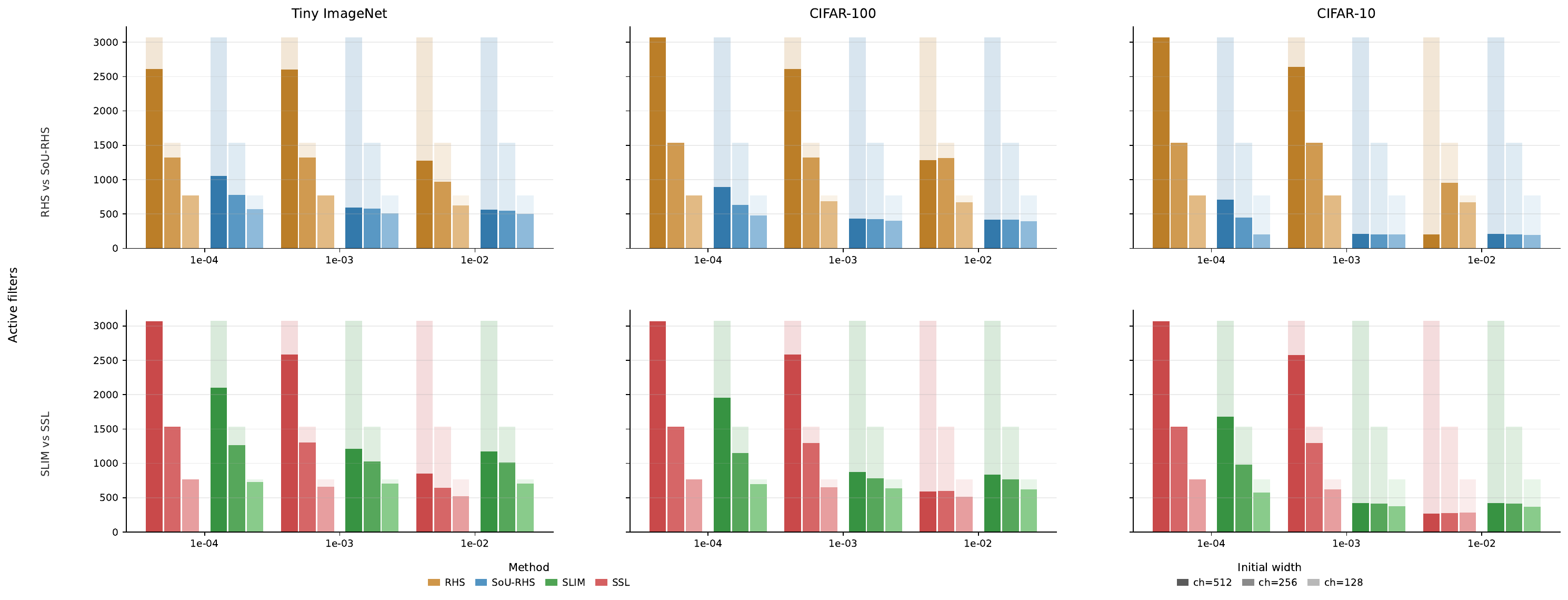}
\caption{
Total number of active filters recovered as the pruning threshold varies, across datasets and starting widths.
Deterministic baselines use regularization strength $\lambda=10^{-3}$.
}
\label{fig:thresholds}
\end{figure}

\paragraph{SoU-SGL stabilizes the recovered active set (Figure~\ref{fig:thresholds}).}

A compact architecture recovery procedure should not be overly sensitive to arbitrary post-processing choices or to the size of the overparameterized model used to initialize it. We therefore first examine the total number of active filters recovered across pruning thresholds and starting widths. Figure~\ref{fig:thresholds} shows that RHS does not provide a stable active set: the number of retained filters changes substantially with the pruning threshold 
and remains sensitive to the starting width.
This behavior reflects the limitation of shrinkage alone as a pruning mechanism: RHS reduces the pruning scores of weak filters, but often leaves them at small nonzero values. As a result, the active set remains sensitive to where the pruning threshold is placed.

SoU-RHS changes this pattern. Across all datasets shown, the recovered number of active filters remains much more stable over the threshold range and across starting widths. 
By driving the pruning scores of weak filters close enough to zero, the SoU-SGL mechanism makes the recovered active set less dependent on the exact pruning cutoff, yielding a sharper active--inactive decision.
The deterministic structured-sparsity baselines also show some stability, but their active-set sizes remain more visibly affected by the pruning threshold and the starting width.

\begin{figure}[h]
    \centering
    \includegraphics[width=\textwidth]{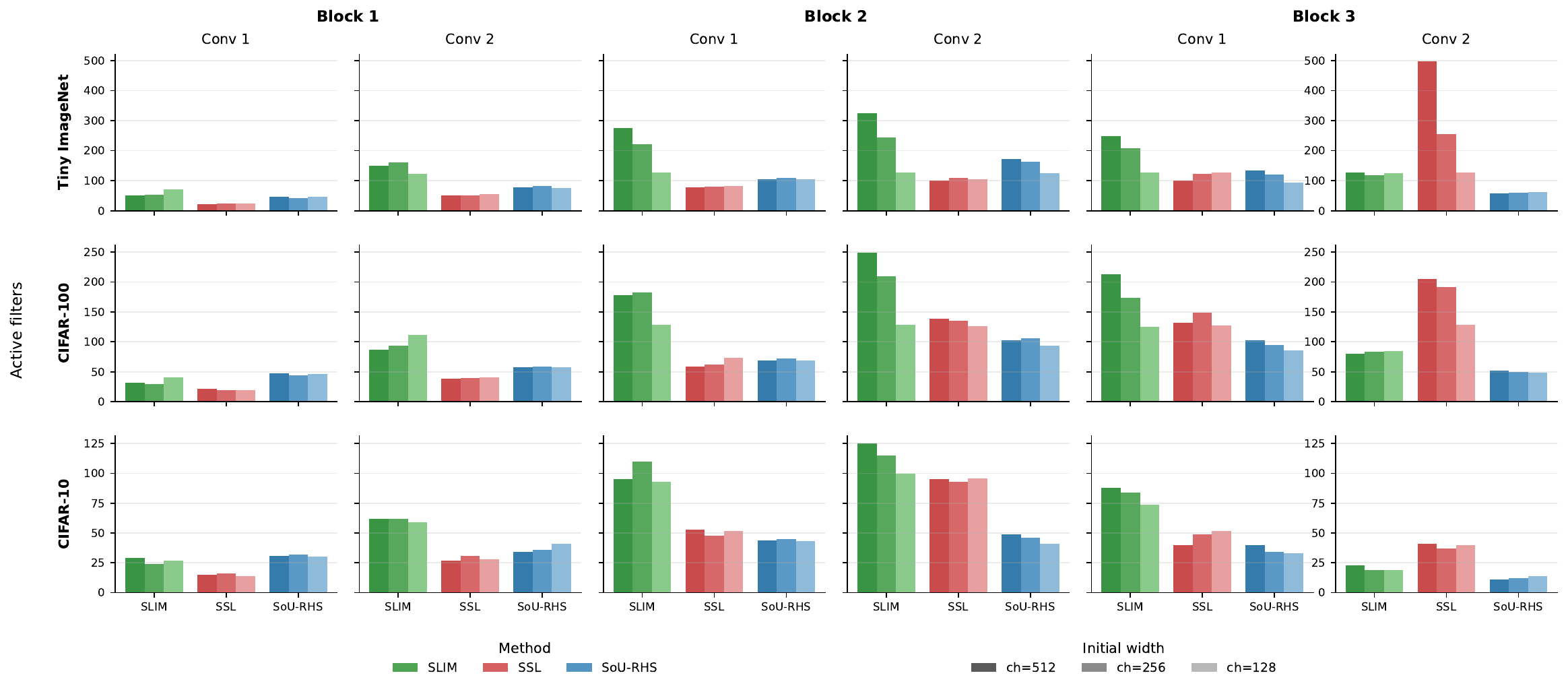}
\caption{
Layer-wise active filters recovered at a representative pruning threshold, across datasets and starting widths.
For SoU-RHS (blue), pruning uses $\delta=10^{-3}$; for SLIM (green) and SSL (red), pruning uses threshold $\delta=10^{-2}$ with regularization strength $\lambda=10^{-3}$.
}
\label{fig:layers}
\end{figure}

\paragraph{Recovered architectures are consistent across starting widths (Figure~\ref{fig:layers}).}
Similar total numbers of active filters can still correspond to different architectures. We therefore examine the number of active filters retained in each convolutional layer at a representative pruning threshold. Figure~\ref{fig:layers} shows that SoU-RHS recovers similar layer-wise widths across starting widths. This consistency strengthens the interpretation of SoU-RHS as recovering a compact architecture that captures the task-relevant structure with fewer filters. The deterministic baselines also recover compact models, but their layer-wise allocations vary more visibly across starting widths.
The recovered SoU-RHS layer-wise allocations also do not mirror the usual hand-designed heuristic of increasing width with depth.

\begin{table}[p]
\vspace{-.7cm}
\caption{
Main comparison across datasets and starting widths.
\#AF$(C)$ denotes the number of active filters after pruning out of $C$ initial filters.
For SLIM and SSL, the column label indicates the regularization strength
$\lambda\in\{10^{-3},10^{-4}\}$.
Predictive metrics for sparsity-inducing methods are evaluated after pruning.
Arrows indicate whether higher or lower values are preferred.
}
\centering
\renewcommand{\arraystretch}{0.94}
\setlength{\tabcolsep}{4pt}

\begin{tabular}{llrccccccc}
\toprule
&&& \multirow{2}{*}{NN} & \multirow{2}{*}{BNN} & SoU-RHS
& \multicolumn{2}{c}{SLIM} & \multicolumn{2}{c}{SSL} \\
\cmidrule(lr){7-8}
\cmidrule(lr){9-10}
Dataset & Width & Stat & & & (ours)
& $10^{-3}$ & $10^{-4}$ & $10^{-3}$ & $10^{-4}$ \\
\midrule

\multirow{12}{*}{\shortstack{Tiny\\ImageNet}} & \multirow{4}{*}{128} & Acc $(\uparrow)$ & 0.356 & 0.321 & \textbf{0.465} & 0.332 & 0.361 & 0.346 & 0.355 \\
 &  & NLL $(\downarrow)$ & 7.932 & 2.900 & \textbf{2.237} & 4.703 & 5.924 & 4.954 & 7.213 \\
 &  & ECE $(\downarrow)$ & 0.510 & \textbf{0.078} & 0.101 & 0.380 & 0.448 & 0.400 & 0.487 \\
 &  & \#AF(768) $(\downarrow)$ & -- & -- & \textbf{509} & 702 & 751 & 522 & 694 \\
\cmidrule(lr){2-10}

 & \multirow{4}{*}{256} & Acc $(\uparrow)$ & 0.417 & 0.294 & \textbf{0.432} & 0.308 & 0.411 & 0.372 & 0.396 \\
 &  & NLL $(\downarrow)$ & 6.871 & 3.073 & \textbf{2.387} & 4.377 & 3.780 & 4.571 & 5.723 \\
 &  & ECE $(\downarrow)$ & 0.453 & 0.084 & \textbf{0.082} & 0.331 & 0.319 & 0.372 & 0.428 \\
 &  & \#AF(1536) $(\downarrow)$ & -- & -- & \textbf{576} & 1010 & 1370 & 645 & 1148 \\
\cmidrule(lr){2-10}

 & \multirow{4}{*}{512} & Acc $(\uparrow)$ & \textbf{0.496} & 0.243 & 0.410 & 0.061 & 0.431 & 0.393 & 0.423 \\
 &  & NLL $(\downarrow)$ & 5.527 & 3.344 & \textbf{2.472} & 7.673 & 3.280 & 4.184 & 5.270 \\
 &  & ECE $(\downarrow)$ & 0.384 & \textbf{0.058} & 0.059 & 0.397 & 0.273 & 0.346 & 0.401 \\
 &  & \#AF(3072) $(\downarrow)$ & -- & -- & \textbf{594} & 1177 & 2018 & 849 & 1433 \\
\midrule

\multirow{12}{*}{CIFAR-100} & \multirow{4}{*}{128} & Acc $(\uparrow)$ & 0.588 & 0.490 & \textbf{0.618} & 0.502 & 0.586 & 0.538 & 0.557 \\
 &  & NLL $(\downarrow)$ & 3.926 & 1.845 & \textbf{1.372} & 2.513 & 2.322 & 3.083 & 3.951 \\
 &  & ECE $(\downarrow)$ & 0.322 & \textbf{0.063} & 0.073 & 0.238 & 0.239 & 0.303 & 0.331 \\
 &  & \#AF(768) $(\downarrow)$ & -- & -- & \textbf{400} & 619 & 695 & 514 & 641 \\
\cmidrule(lr){2-10}

 & \multirow{4}{*}{256} & Acc $(\uparrow)$ & \textbf{0.652} & 0.466 & 0.627 & 0.490 & 0.602 & 0.538 & 0.618 \\
 &  & NLL $(\downarrow)$ & 2.941 & 1.965 & \textbf{1.330} & 2.614 & 1.974 & 3.074 & 2.869 \\
 &  & ECE $(\downarrow)$ & 0.264 & 0.073 & \textbf{0.065} & 0.237 & 0.199 & 0.305 & 0.273 \\
 &  & \#AF(1536) $(\downarrow)$ & -- & -- & \textbf{426} & 770 & 1099 & 595 & 928 \\
\cmidrule(lr){2-10}

 & \multirow{4}{*}{512} & Acc $(\uparrow)$ & \textbf{0.671} & 0.418 & 0.631 & 0.301 & 0.630 & 0.553 & 0.621 \\
 &  & NLL $(\downarrow)$ & 2.543 & 2.161 & \textbf{1.318} & 3.104 & 1.772 & 2.870 & 2.254 \\
 &  & ECE $(\downarrow)$ & 0.242 & 0.077 & 0.067 & \textbf{0.024} & 0.176 & 0.290 & 0.234 \\
 &  & \#AF(3072) $(\downarrow)$ & -- & -- & \textbf{430} & 839 & 1265 & 594 & 864 \\
\midrule

\multirow{12}{*}{CIFAR-10} & \multirow{4}{*}{128} & Acc $(\uparrow)$ & \textbf{0.877} & 0.816 & 0.869 & 0.498 & 0.812 & 0.817 & 0.853 \\
 &  & NLL $(\downarrow)$ & 1.011 & 0.547 & \textbf{0.392} & 4.819 & 0.969 & 0.908 & 1.039 \\
 &  & ECE $(\downarrow)$ & 0.102 & 0.024 & \textbf{0.019} & 0.381 & 0.120 & 0.122 & 0.113 \\
 &  & \#AF(768) $(\downarrow)$ & -- & -- & \textbf{202} & 372 & 526 & 282 & 445 \\
\cmidrule(lr){2-10}

 & \multirow{4}{*}{256} & Acc $(\uparrow)$ & \textbf{0.895} & 0.799 & 0.870 & 0.572 & 0.857 & 0.829 & 0.860 \\
 &  & NLL $(\downarrow)$ & 0.820 & 0.607 & \textbf{0.388} & 3.155 & 0.689 & 0.811 & 0.956 \\
 &  & ECE $(\downarrow)$ & 0.085 & 0.025 & \textbf{0.015} & 0.310 & 0.088 & 0.114 & 0.106 \\
 &  & \#AF(1536) $(\downarrow)$ & -- & -- & \textbf{205} & 414 & 599 & 274 & 467 \\
\cmidrule(lr){2-10}

 & \multirow{4}{*}{512} & Acc $(\uparrow)$ & \textbf{0.899} & 0.785 & 0.873 & 0.766 & 0.844 & 0.830 & 0.858 \\
 &  & NLL $(\downarrow)$ & 0.748 & 0.631 & \textbf{0.378} & 0.933 & 0.744 & 0.871 & 0.900 \\
 &  & ECE $(\downarrow)$ & 0.080 & 0.038 & \textbf{0.020} & 0.117 & 0.097 & 0.114 & 0.107 \\
 &  & \#AF(3072) $(\downarrow)$ & -- & -- & \textbf{209} & 422 & 639 & 271 & 471 \\
\midrule

\multirow{12}{*}{SVHN} & \multirow{4}{*}{128} & Acc $(\uparrow)$ & \textbf{0.952} & 0.936 & \textbf{0.952} & 0.861 & 0.948 & 0.934 & 0.941 \\
 &  & NLL $(\downarrow)$ & 0.386 & 0.225 & \textbf{0.176} & 0.515 & 0.265 & 0.315 & 0.393 \\
 &  & ECE $(\downarrow)$ & 0.038 & 0.016 & \textbf{0.009} & 0.050 & 0.030 & 0.035 & 0.042 \\
 &  & \#AF(768) $(\downarrow)$ & -- & -- & \textbf{115} & 292 & 402 & 190 & 313 \\
\cmidrule(lr){2-10}

 & \multirow{4}{*}{256} & Acc $(\uparrow)$ & \textbf{0.956} & 0.934 & 0.950 & 0.146 & 0.932 & 0.932 & 0.941 \\
 &  & NLL $(\downarrow)$ & 0.368 & 0.245 & \textbf{0.178} & 4.488 & 0.273 & 0.330 & 0.396 \\
 &  & ECE $(\downarrow)$ & 0.035 & 0.032 & \textbf{0.008} & 0.686 & 0.024 & 0.038 & 0.042 \\
 &  & \#AF(1536) $(\downarrow)$ & -- & -- & \textbf{117} & 302 & 429 & 180 & 310 \\
\cmidrule(lr){2-10}

 & \multirow{4}{*}{512} & Acc $(\uparrow)$ & \textbf{0.955} & 0.926 & 0.950 & 0.184 & 0.928 & 0.930 & 0.940 \\
 &  & NLL $(\downarrow)$ & 0.374 & 0.255 & \textbf{0.178} & 2.725 & 0.262 & 0.331 & 0.377 \\
 &  & ECE $(\downarrow)$ & 0.035 & 0.019 & \textbf{0.009} & 0.090 & 0.010 & 0.036 & 0.042 \\
 &  & \#AF(3072) $(\downarrow)$ & -- & -- & \textbf{118} & 301 & 442 & 182 & 310 \\

\bottomrule
\end{tabular}
\label{tab:all-datasets-no-ap-no-rhs}
\end{table}

\paragraph{SoU-RHS achieves compactness with competitive predictive performance (Table~\ref{tab:all-datasets-no-ap-no-rhs}).}
SoU-RHS recovers the smallest architectures in terms of active filters in all settings, as reported in the main quantitative comparison in Table~\ref{tab:all-datasets-no-ap-no-rhs}. Moreover, its active-filter counts vary less across starting widths than those of the deterministic structured-sparsity baselines.

This compactness should be interpreted together with the underlying mechanism that produces it. For the deterministic baselines, SLIM and SSL, the degree of compression is controlled by the explicit regularization parameter $\lambda$, inducing a trade-off: weaker regularization preserves accuracy by retaining larger architectures, whereas stronger regularization increases compression but often degrades prediction. SoU-RHS does not introduce an explicit sparsity-strength parameter; once the local confidence is set near zero, the amount of sparsity is determined by the fitted global-local posterior.

This makes the compactness of SoU-RHS more notable. Even the more aggressively regularized deterministic baselines retain more active filters, while already paying a substantial cost in predictive quality. 

Despite this stronger compression, SoU-RHS remains competitive in accuracy with the dense NN.  
On the more challenging datasets, this comparison is especially favorable at smaller starting widths: SoU-RHS outperforms the dense NN on Tiny-ImageNet for widths 128 and 256, and on CIFAR-100 for width 128. 
As the dense NN becomes wider, its accuracy tends to increase, suggesting that it benefits more directly from the additional overparameterization.

SoU-RHS also shows a more consistent advantage over the baselines in probabilistic predictive performance. 
It obtains the best NLL in every configuration and substantially improves calibration relative to the deterministic baselines. The Gaussian BNN sometimes attains lower ECE, but these cases coincide with much lower accuracy. Thus, the proposed Bayesian structured-sparsity formulation is not only useful for recovering compact architectures, but also preserves the probabilistic advantages expected from Bayesian inference.

\paragraph{Predictive performance under pruning.} Table~\ref{tab:rhs-sou-threshold-size-matched} complements the structural analysis in Figure~\ref{fig:thresholds} by examining the predictive consequences of pruning. At common thresholds, both methods generally 
retain much of their pre-pruning accuracy, 
except for the larger Tiny-ImageNet models at \(\delta=10^{-2}\). However, SoU-RHS achieves this with substantially fewer active filters, while RHS typically undergoes much milder pruning. To assess whether this difference merely reflects threshold calibration, we also adjust the RHS threshold to match the number of filters retained by SoU-RHS at \(\delta=10^{-3}\), using filter counts alone. At matched sizes, RHS remains competitive on CIFAR-10. 
On CIFAR-100 and Tiny-ImageNet, however, the more aggressive pruning leads to severe accuracy degradation for RHS in most configurations. 
Overall, SoU-RHS achieves stronger, more threshold-stable pruning while retaining higher accuracy than RHS at matched filter counts.

\begin{table}[h]
\centering
\caption{Predictive performance and active filters across pruning thresholds for SoU-RHS and RHS. The size-matched column evaluates RHS at the active-filter count of SoU-RHS for $\delta=10^{-3}$. \#AF($C$) denotes the number of active filters after pruning out of $C$ initial filters. Bold indicates the smaller active-filter count at each common threshold. Arrows indicate whether higher or lower values are preferred.}
\label{tab:rhs-sou-threshold-size-matched}
\small
\setlength{\tabcolsep}{4pt}
\begin{tabular}{llrcccccc@{\hspace{8pt}}c}
\toprule
 & & & \multicolumn{2}{c}{$\delta=10^{-4}$} & \multicolumn{2}{c}{$\delta=10^{-3}$} & \multicolumn{2}{c}{$\delta=10^{-2}$} & \multicolumn{1}{c}{Size-matched} \\
\cmidrule(lr){4-5}
\cmidrule(lr){6-7}
\cmidrule(lr){8-9}
\cmidrule(l){10-10}
Dataset & Width & Stat & RHS & SoU-RHS & RHS & SoU-RHS & RHS & SoU-RHS & \multicolumn{1}{c}{RHS} \\
\midrule
\multirow{6}{*}{\shortstack{Tiny\\ImageNet}} & \multirow{2}{*}{128} & Acc $(\uparrow)$ & 0.466 & 0.464 & 0.466 & 0.464 & 0.466 & 0.423 & 0.151 \\
 &  & \#AF(768) $(\downarrow)$ & 768 & \textbf{574} & 768 & \textbf{509} & 624 & \textbf{504} & 509 \\
\addlinespace[3pt]
 & \multirow{2}{*}{256} & Acc $(\uparrow)$ & 0.471 & 0.471 & 0.436 & 0.431 & 0.007 & 0.007 & 0.006 \\
 &  & \#AF(1536) $(\downarrow)$ & 1323 & \textbf{776} & 1320 & \textbf{576} & 973 & \textbf{547} & 576 \\
\addlinespace[3pt]
 & \multirow{2}{*}{512} & Acc $(\uparrow)$ & 0.471 & 0.472 & 0.404 & 0.409 & 0.006 & 0.004 & 0.005 \\
 &  & \#AF(3072) $(\downarrow)$ & 2611 & \textbf{1051} & 2603 & \textbf{594} & 1273 & \textbf{566} & 594 \\
\midrule
\multirow{6}{*}{CIFAR-100} & \multirow{2}{*}{128} & Acc $(\uparrow)$ & 0.616 & 0.620 & 0.615 & 0.618 & 0.581 & 0.545 & 0.024 \\
 &  & \#AF(768) $(\downarrow)$ & 768 & \textbf{481} & 682 & \textbf{400} & 673 & \textbf{392} & 400 \\
\addlinespace[3pt]
 & \multirow{2}{*}{256} & Acc $(\uparrow)$ & 0.625 & 0.628 & 0.625 & 0.628 & 0.586 & 0.608 & 0.005 \\
 &  & \#AF(1536) $(\downarrow)$ & 1536 & \textbf{633} & 1323 & \textbf{426} & 1316 & \textbf{421} & 426 \\
\addlinespace[3pt]
 & \multirow{2}{*}{512} & Acc $(\uparrow)$ & 0.628 & 0.631 & 0.629 & 0.631 & 0.613 & 0.570 & 0.603 \\
 &  & \#AF(3072) $(\downarrow)$ & 3072 & \textbf{890} & 2607 & \textbf{430} & 1286 & \textbf{421} & 430 \\
\midrule
\multirow{6}{*}{CIFAR-10} & \multirow{2}{*}{128} & Acc $(\uparrow)$ & 0.866 & 0.870 & 0.866 & 0.870 & 0.864 & 0.868 & 0.860 \\
 &  & \#AF(768) $(\downarrow)$ & 768 & \textbf{203} & 768 & \textbf{202} & 667 & \textbf{199} & 202 \\
\addlinespace[3pt]
 & \multirow{2}{*}{256} & Acc $(\uparrow)$ & 0.865 & 0.870 & 0.865 & 0.870 & 0.866 & 0.868 & 0.848 \\
 &  & \#AF(1536) $(\downarrow)$ & 1536 & \textbf{449} & 1536 & \textbf{205} & 952 & \textbf{203} & 205 \\
\addlinespace[3pt]
 & \multirow{2}{*}{512} & Acc $(\uparrow)$ & 0.869 & 0.873 & 0.869 & 0.873 & 0.863 & 0.871 & 0.869 \\
 &  & \#AF(3072) $(\downarrow)$ & 3072 & \textbf{711} & 2643 & \textbf{209} & \textbf{205} & 207 & 209 \\
\bottomrule
\end{tabular}
\vspace{2pt}
\parbox{\linewidth}{\footnotesize\raggedright
\emph{Size-matched RHS thresholds} $\delta^\star$ (widths 128, 256, 512):
Tiny-ImageNet: $(0.07288,\,0.11064,\,0.03329)$;\enspace
CIFAR-100: $(0.05239,\,0.05915,\,0.01385)$;\enspace
CIFAR-10: $(0.04809,\,0.05396,\,0.005238)$.
}
\end{table}

\paragraph{Retraining reveals architecture--inference interaction (Table~\ref{tab:compact-from-scratch-all-datasets}).}
\citet{liu2019rethinking} showed that structured-pruning benefits can arise from the recovered architecture itself, with training from scratch matching or exceeding fine-tuning in their experiments. Motivated by this, we ask whether the compact architecture recovered by SoU-RHS remains effective across different inference procedures.
We therefore retrain all methods from scratch on the compact architecture obtained from the width-512 SoU-RHS fit. 
The deterministic methods deteriorate substantially relative to their overparameterized counterparts, with lower accuracy and worse NLL and calibration 
in most cases.
In contrast, SoU-RHS remains close to its original performance, with RHS achieving similar results. 
This indicates that the recovered architecture is not simply a method-agnostic compact network: its effectiveness depends on the inference procedure used on it.

\begin{table}[h]
\caption{
Model performance after training from scratch on the compact architecture recovered by SoU-RHS from the width-512 starting model.
For each dataset, all methods use the same recovered architecture. Arrows indicate whether higher or lower values are preferred.
}
\centering
\setlength{\tabcolsep}{3.4pt}
\begin{tabular}{lrcccccccc}
\toprule
 & & \multirow{2}{*}{NN} & \multirow{2}{*}{BNN} & \multirow{2}{*}{SoU-RHS} & \multirow{2}{*}{RHS} & \multicolumn{2}{c}{SLIM} & \multicolumn{2}{c}{SSL} \\
\cmidrule(lr){7-8}\cmidrule(lr){9-10}
Dataset & Stat & & & & & $10^{-3}$ & $10^{-4}$ & $10^{-3}$ & $10^{-4}$ \\
\midrule
\multirow{3}{*}{Tiny ImageNet} & Acc $(\uparrow)$ & 0.332 & 0.333 & \textbf{0.464} & 0.459 & 0.301 & 0.328 & 0.309 & 0.314 \\
 & NLL $(\downarrow)$ & 8.604 & 2.851 & 2.260 & \textbf{2.253} & 5.112 & 7.306 & 5.336 & 7.664 \\
 & ECE $(\downarrow)$ & 0.529 & \textbf{0.073} & 0.104 & 0.099 & 0.410 & 0.496 & 0.419 & 0.514 \\
\midrule
\multirow{3}{*}{CIFAR-100} & Acc $(\uparrow)$ & 0.523 & 0.535 & \textbf{0.613} & 0.610 & 0.502 & 0.526 & 0.506 & 0.515 \\
 & NLL $(\downarrow)$ & 5.002 & 1.685 & \textbf{1.389} & 1.397 & 2.776 & 3.397 & 3.600 & 4.844 \\
 & ECE $(\downarrow)$ & 0.379 & 0.072 & 0.073 & \textbf{0.070} & 0.279 & 0.325 & 0.339 & 0.378 \\
\midrule
\multirow{3}{*}{CIFAR-10} & Acc $(\uparrow)$ & 0.809 & 0.822 & 0.849 & \textbf{0.852} & 0.795 & 0.799 & 0.797 & 0.801 \\
 & NLL $(\downarrow)$ & 1.666 & 0.516 & 0.448 & \textbf{0.443} & 1.101 & 1.530 & 1.130 & 1.540 \\
 & ECE $(\downarrow)$ & 0.160 & 0.016 & \textbf{0.006} & 0.014 & 0.143 & 0.163 & 0.145 & 0.160 \\
\midrule
\multirow{3}{*}{SVHN} & Acc $(\uparrow)$ & 0.908 & \textbf{0.947} & 0.946 & \textbf{0.947} & 0.914 & 0.913 & 0.918 & 0.907 \\
 & NLL $(\downarrow)$ & 0.734 & 0.192 & 0.191 & \textbf{0.187} & 0.496 & 0.649 & 0.362 & 0.669 \\
 & ECE $(\downarrow)$ & 0.073 & 0.020 & \textbf{0.007} & 0.008 & 0.056 & 0.068 & 0.041 & 0.070 \\
\bottomrule
\end{tabular}
\label{tab:compact-from-scratch-all-datasets}
\end{table}

The Gaussian BNN makes this interaction particularly clear. In the overparameterized models of Table~\ref{tab:all-datasets-no-ap-no-rhs}, its predictive performance deteriorates as the starting width increases and remains well below that of the deterministic NN. On the recovered compact architecture, however, the same BNN improves substantially and becomes one of the strongest models, outperforming the deterministic NN across all four datasets. This suggests that architectures favorable to deterministic training need not be equally favorable to Bayesian inference, and that architecture recovery and inference should not be viewed as entirely separate design choices.

\section{Limitations\label{sec:lim}}

Our evaluation is restricted to VGG-style convolutional networks and datasets no larger than Tiny-ImageNet. 
While this controlled setting is well suited for studying the stability of
architecture recovery across starting widths and pruning thresholds, it does not
establish how SoU-RHS scales to more complex architectures or larger-scale
datasets.

SoU-RHS incurs a higher training cost than the deterministic baselines, mainly
because we use longer training schedules for the Bayesian models. 
Consistent with the behavior observed for SoU-SGL, sparsity can continue to increase after
predictive accuracy has largely stabilized. Since our primary goal is compact
architecture recovery, we use long training runs to allow the shrinkage scales
and the recovered active set to stabilize. 
Improving the efficiency of the training protocol remains a practical avenue for improvement.

Finally, architectural compactness does not directly translate into predictive cost for Bayesian models, since posterior predictive averaging requires multiple forward passes. Appendix~\ref{sec:post_sampling} quantifies this trade-off through the predictive FLOPs ratio, which accounts jointly for architectural compression and the number of posterior samples.

\section{Conclusion\label{sec:conc}}

We introduced SoU-RHS, a Bayesian structured-sparsity procedure that combines regularized horseshoe shrinkage with SoU-SGL for compact architecture recovery. 
By selectively treating the filter-specific local scales, SoU-RHS produces a clearer separation between active and inactive filters, turning horseshoe shrinkage into a more decisive architecture-selection mechanism.
The central empirical finding is stability: SoU-RHS recovers compact architectures with similar total active-set sizes and layer-wise allocations across overparameterized starting widths and under a reasonable range of pruning thresholds.

Across the image-classification datasets considered, the recovered architectures are substantially smaller than those obtained by deterministic baselines while maintaining competitive accuracy and improving probabilistic performance, as measured by NLL and calibration. 
Retraining these same architectures further suggests that they are better suited to Bayesian models than to deterministic ones.

Although we applied the method here to CNNs, the procedure is more general and can be naturally extended to other architectures by placing a horseshoe prior on their structural components and strengthening the induced shrinkage mechanism with SoU-SGL.

\bibliography{biblio}
\bibliographystyle{tmlr}

\clearpage
\appendix

\section{Posterior Sampling and Predictive Cost}\label{sec:post_sampling}

Bayesian prediction introduces an additional inference cost through posterior predictive averaging, which requires multiple forward passes. To quantify this cost, we define the architectural FLOPs ratio as the cost of one forward pass through the pruned architecture divided by that of the corresponding dense starting architecture. For \(M\) posterior samples, we define the predictive FLOPs ratio (PFR) as
\[
\mathrm{PFR}(M)
=
M \times \mathrm{FLOPs\ ratio}.
\]
Table~\ref{tab:sou-rhs-pruned-mc-sweep-vs-nn-256-absolute} makes this cost--quality trade-off explicit by comparing the pruned SoU-RHS predictor with the dense NN baseline as the number of Monte Carlo samples varies.

\begin{table}[h]
\caption{
Pruned SoU-RHS performance relative to the dense NN baseline for the width-256 architecture as the number of Monte Carlo samples $M$ varies.
For accuracy, $\Delta$ is SoU-RHS minus NN; for NLL and ECE, $\Delta$ is NN minus SoU-RHS, so positive values always indicate improvement over the NN. 
Entries report mean differences across \(100\) independent stochastic evaluations, with standard deviations in parentheses. 
Arrows indicate whether higher or lower values are preferred.
}
\centering
\setlength{\tabcolsep}{4pt}
\begin{tabular}{crccccc}
\toprule
Dataset & Stat & $M=1$ & $M=2$ & $M=3$ & $M=5$ & $M=10$ \\
\midrule
 & $\Delta$Acc $(\uparrow)$ & $-$0.011 (0.002) & +0.001 (0.002) & +0.006 (0.002) & +0.009 (0.002) & +0.012 (0.001) \\
 Tiny & $\Delta$NLL $(\uparrow)$ & +4.377 (0.005) & +4.428 (0.003) & +4.447 (0.002) & +4.462 (0.002) & +4.474 (0.001) \\
 ImageNet & $\Delta$ECE $(\uparrow)$ & +0.417 (0.002) & +0.396 (0.002) & +0.388 (0.002) & +0.382 (0.002) & +0.378 (0.001) \\
 & PFR$(M)$ $(\downarrow)$ & 0.111 & 0.221 & 0.332 & 0.553 & 1.106 \\
\midrule
\multirow{4}{*}{CIFAR-100} & $\Delta$Acc $(\uparrow)$ & $-$0.074 (0.003) & $-$0.051 (0.002) & $-$0.042 (0.002) & $-$0.034 (0.002) & $-$0.029 (0.002) \\
 & $\Delta$NLL $(\uparrow)$ & +1.409 (0.006) & +1.508 (0.005) & +1.542 (0.004) & +1.569 (0.003) & +1.591 (0.002) \\
 & $\Delta$ECE $(\uparrow)$ & +0.232 (0.003) & +0.244 (0.002) & +0.231 (0.002) & +0.217 (0.002) & +0.207 (0.002) \\
 & PFR$(M)$ $(\downarrow)$ & 0.062 & 0.124 & 0.185 & 0.309 & 0.618 \\
\midrule
\multirow{4}{*}{CIFAR-10} & $\Delta$Acc $(\uparrow)$ & $-$0.058 (0.002) & $-$0.042 (0.002) & $-$0.036 (0.002) & $-$0.032 (0.001) & $-$0.028 (0.001) \\
 & $\Delta$NLL $(\uparrow)$ & +0.304 (0.006) & +0.370 (0.004) & +0.392 (0.003) & +0.408 (0.002) & +0.421 (0.002) \\
 & $\Delta$ECE $(\uparrow)$ & +0.031 (0.002) & +0.065 (0.002) & +0.072 (0.002) & +0.075 (0.002) & +0.073 (0.001) \\
 & PFR$(M)$ $(\downarrow)$ & 0.021 & 0.043 & 0.064 & 0.107 & 0.214 \\
\midrule
\multirow{4}{*}{SVHN} & $\Delta$Acc $(\uparrow)$ & $-$0.017 (0.001) & $-$0.011 (0.001) & $-$0.009 (0.001) & $-$0.008 (0.001) & $-$0.007 (0.000) \\
 & $\Delta$NLL $(\uparrow)$ & +0.145 (0.002) & +0.167 (0.001) & +0.175 (0.001) & +0.181 (0.001) & +0.185 (0.001) \\
 & $\Delta$ECE $(\uparrow)$ & +0.019 (0.001) & +0.028 (0.001) & +0.029 (0.001) & +0.029 (0.001) & +0.028 (0.001) \\
 & PFR$(M)$ $(\downarrow)$ & 0.005 & 0.010 & 0.016 & 0.026 & 0.052 \\
\bottomrule
\end{tabular}
\label{tab:sou-rhs-pruned-mc-sweep-vs-nn-256-absolute}
\end{table}

The competitive accuracy and uncertainty quality of SoU-RHS already materialize with few posterior samples, with substantial probabilistic gains appearing with a single posterior sample. At \(M=1\), SoU-RHS improves both NLL and ECE over the dense NN on every dataset, despite requiring only one forward pass through the pruned architecture.
Increasing \(M\) then provides the expected trade-off: accuracy improves as more posterior samples are averaged, but the predictive FLOPs ratio increases linearly. Thus, \(M\) acts as a deployment knob, allowing one to trade additional computation for improved predictive accuracy while retaining the probabilistic gains observed even at small \(M\).

\end{document}